# Towards Transliteration between Sindhi Scripts from Devanagari to Perso-Arabic

Shivani Singh Rathore[1], Bharti Nathani[1,2], Nisheeth Joshi[1,2], Pragya Katyayan[1,2]
[1]Department of Computer Science, Banasthali Vidyapith, Rajasthan, India
[2]Centre for Artificial Intelligence, Banasthali Vidyapith, Rajasthan
ssrathore5995@gmail.com, nbharti@banasthali.in, nisheeth.joshi@rediffmail.com, pragya.katyayan@outlook.com

Chander Prakash Dadlani[3]
[3]Department of Sindhi
Samrat Prithviraj Chauhan Government PG College
Rajasthan, India
chander.dadlani@gmail.com

*Abstract*—In this paper, we have shown a script conversion (transliteration) technique that converts Sindhi text in the Devanagari script to the Perso-Arabic script. We showed this by incorporating a hybrid approach where some part of the text is converted using a rule base and in case an ambiguity arises then a probabilistic model is used to resolve the same. Using this approach, the system achieved an overall accuracy of 99.64%.

## I. Introduction

Machine transliteration is the process of converting text written in one script (e.g., Roman) into a corresponding text written in another script (e.g., Devanagari). It aims to preserve the pronunciation of the original text while mapping it to a new script. Transliteration is commonly used in the field of NLP for tasks such as information retrieval and machine translation, where text written in different scripts needs to be processed and analyzed. Machine transliteration can be accomplished using various techniques such as rule-based methods, statistical models, or a combination of both. The goal of machine transliteration is to provide a high-quality, automated way of converting text written in one script into another, which can enable better processing and analysis of multilingual text.

The need for machine transliteration arises from the fact that text written in different scripts can be difficult to process and analyze for natural language processing (NLP) tasks. This is particularly relevant for low-resource languages, where resources and data in one script may be more abundant than in another script. By transliterating text written in one script into another, machine transliteration can help to overcome these barriers and enable better processing and analysis of multilingual text.

Additionally, machine transliteration can help to support multilingual information retrieval and machine translation by providing a bridge between different scripts. In some cases, it can also help to preserve the cultural heritage of languages by enabling the representation of text written in less widely used scripts in a more widely used script.

Overall, machine transliteration is an important tool for NLP tasks, as it can help to overcome barriers in processing and analyzing multilingual text, support multilingual information retrieval and machine translation, and preserve cultural heritage.

There are mainly two types of machine transliteration:

Phonetic transliteration: This type of transliteration aims to preserve the pronunciation of the original text as closely as possible, regardless of the exact spelling in the target script. For example, converting the English name "Ashok Singh" to the Devanagari script as "अशोक सिंह".

Orthographic transliteration: This type of transliteration aims to preserve the spelling of the original text as closely as possible, regardless of the exact pronunciation in the target script. For example, converting the Urdu word "شہر" to the Devanagari script as "शहर ".

Both phonetic and orthographic transliteration have their advantages and limitations, and the choice between them may depend on the specific NLP task and the desired output. Additionally, some systems may use a combination of both approaches, or employ additional methods such as rule-based methods, statistical models, or machine learning algorithms to improve the quality of the transliteration.

In this paper we have used a mix of rule-based and statistical learning for automatic transliteration of Sindhi text in Devanagari script into Perso-Arabic script.

## II. Literature Survey

Leghari and Rahman [1] have presented transliteration models for Sindhi-Arabic and Sindhi-Devanagari based on roman script as intermediary. They have proposed an algorithm that transliterates between Sindhi-Arabic and Sindhi-Devanagari. Finch et al. [2] have examined the worth of seq2seq model for transliteration task. They have found that on including a neural score as feature with phrase-based statistical transliteration system, the performance is enhanced. Kunchukuttan et al. [3] presented a transliteration system for all major Indian languages (306 pairs) on an online platform named- Bramhi-Net. System involves English, 13 Indo-aryan languages and 4 Dravidian languages.

Androutsopoulos [4] introduced 'Greeklish' which is a transliteration practice and discourse for computerised digraphia. Rosca and Breuel [5] have developed seq2seq neural network models for transliteration and have reported close to state-of-the-art results. They open-sourced Arabic to English dataset and transliteration models also. Guellil et al. [6] proposed a technique to apply character based neural transliteration for Arabizi to Arabic scripts. They also constructed Arabizi corpus for the task.

Alam and Hussain [7] have transformed transliteration of Roman Urdu to Urdu in a seq2seq problem. They created a corpus for the above language pair along with a neural transliteration model. Kundu et al. [8] proposed a deep learning-based language independent machine transliteration system for named entities using RNN and convolutional seq2seq based NMT model. Merhav and Ash [9] analyzed the challenges affecting the development of transliteration system for named entities in multiple languages. They evaluated encoder-decoder RNN methods as well as non-sequential transformer techniques.

Ahmadi [10] presented a rule-based approach for Sorani to Kurdish transliteration. They identified characters by eliminating ambiguities and mapped it to target language script. Hossain et al. [11] developed a system based on Levenshtein distance that works better than other systems by spell-checking the transliterated word. Shazal et al. [12] presented a unified model for detecting and transliterating Arabizi to code-mixed output by using a deep learning seq2seq model.

Mansurov and Mansurov [13] introduced transliteration of Uzbek words from Cyrillic to Latin script and vice versa using Decision tree classifier to learn the character alignment of words. Khare et al. [14] proposed a new technique of pre-training transfer learning models using huge speech data in high resourced language and its text transliterated in low-resource language. Al-Jarf [15] explored the transliteration of geminated Arabic names to English on social media and their anomalies. They took 406 English samples of Arabic names with geminates from Facebook and used them to study correct transliteration of double consonants. Madhani et al. [16] introduced largest open-source dataset for transliteration in 21 Indian languages having 26 million transliteration pairs.

### III. CHALLENGES IN TRANSLITERATION OF DEVANAGARI SCRIPT INTO PERSO-ARABIC SCRIPT

There are several challenges in transliteration of Devanagari script into the Perso-Arabic script:

- **Complex Scripts:** Both Devanagari and Perso-Arabic are complex scripts with a large number of characters and ligatures, which can make transliteration difficult.

- **Multiple Pronunciations:** There may be multiple ways to pronounce a word in Devanagari, and the chosen transliteration may depend on the target language and dialect.

- Ambiguity: Some Devanagari characters have multiple meanings and can be transliterated differently depending on context.

- **Lack of Standardization:** There is a lack of standardization in transliteration between Devanagari and Perso-Arabic, which can lead to inconsistencies and difficulties in information retrieval and machine translation tasks.

- **Cross-Script Differences:** The differences in character shapes, writing direction, and word order between Devanagari and Perso-Arabic can make transliteration challenging.:

### IV. PROPOSED METHODOLOGY

To implement a Sindhi-Devnagari to Sindhi-PersoArabic transliteration system, have used a hybrid approach which is a mix of rule-based and machine learning approaches. As a first step to our system, we have extracted phonemes from the input text and then created a rule base for mapping of Sindhi-Devanagari characters into Sindhi-PersoArabic characters. Thus, we created rules for three types of Sindhi-Devanagari characters. These were:

#### A. Consonants

There are 43 consonants in Sindhi-Devanagari which are ten more than Hindi. These 6 characters having nuktas in them (ख़, ग़, ज़, ड़, ढ़, फ़) and four characters having diacritics (ग॒, ज॒, ड॒, ह॒). Thus, mapping for all these characters were created. A suggestive mapping is shown in table 1.

TABLE I. SUGGESTIVE MAPPING BETWEEN SINDHI PERSO-ARABIC AND SINDHI-DEVANAGARI CHARACTERS

| Sindhi-Devanagari Character | Sindhi-PersoArabic Character |
|---|---|
| ڪ | क |
| ک | ख |
| گ | ग |
| ڳ | ग॒ |

#### B. Vowels

There are 11 vowel characters in Sindhi-Devanagari. These are (अ, आ, इ, ई, उ, ऊ, ए, ऐ, ओ, औ, अं).

## C. Vowel Symbols

Among the 13 vowel characters 12 of these can be transformed in vowel symbols(diacritics/matras). These are: (ा,ि,ी,ु,ू,ॆ,ॊ,े,ो,ॅ,ॉ,ः). Moreover, these vowels and vowel symbols can be classified into two categories viz short vowel (ह्रस्व स्वर) and long vowel (दीर्घ स्वर). Further these vowels and vowel symbols share the same set of characters in Sindhi Perso-Arabic script. these are shown in table 2.

TABLE II. MAPPING BETWEEN SINDHI-DEVANAGARI VOWEL AND VOWEL SYMBOLS AND SINDHI PERSO-ARABIC CHARACTERS

| Vowels | Vowel Symbols | Perso-Arabic Characters |
|---|---|---|
| अ | - | ا |
| आ | ा | آ |
| इ | ि | ا |
| ई | ी | ئي |
| उ | ु | أ |
| ऊ | ू | أو |
| ए | े | اي |
| ऐ | ै | ائي |
| ओ | ो | ا |
| औ | ौ | ائو |
| अं | ं | ن |

The phonification algorithm works as follows:

1. Input Sindhi Devanagari text.
2. Identify Vowels, Vowel Symbols and Consonants.
3. Identify Consonant-Vowel Symbol Combination and consider them as one phoneme.
4. Identify Vowels followed by Consonants and consider them as two separate phonemes.
5. Consider Vowel surrounded by two consonants as separate phonemes.
6. Transliterate characters of each phoneme into Sindhi Perso-Arabic.

This algorithm would mostly generate correct transliterations. Some of the characters having multiple mapping were not resolved, thus a disambiguation module was used. For this we created a knowledge base of phonemes of both the scripts and generated bigram and trigram probabilities for them using equation 1 and 2 respectively. Where $C_{i-2}$, $C_{i-1}$ and $C_i$ were the characters in Devanagari script; $C_i$ being the character which has more than one mapping and $C_{i-2}$, $C_{i-1}$ are the characters before the character where the ambiguity occurs.

$$P(C_i|C_{i-1}) = \frac{Freq(C_{i-1},C_i)}{Freq(C_{i-1})} \quad (1)$$

$$P(C_i|C_{i-2}C_{i-1}) = \frac{Freq(C_{i-2},C_{i-1},C_i)}{Freq(C_{i-2},C_{i-1})} \quad (2)$$

Moreover, we also calculated the probability of Devanagari and with Perso-Arabic character as shown in equation 3. Finally, we applied Hidden Markov Model (HMM) for applying the disambiguation. Equations for bigram HMM is shown in equations 4.

$$P(C_i|B_i) = \frac{Freq(B_i,C_i)}{Freq(B_i)} \quad (3)$$

$$P(B_i|C_i) = P(C_i|B_i) \times P(C_i|C_{i-1}) \times P(C_{i+1}|C_i) \quad (4)$$

The working of the system is shown in figure 1 where an input sentence is sent to the phonification module which extracts the phonemes and maps the characters of two scripts. In case if there is more than one mapping found then disambiguation module is called which applied HMM procedure to identify correct mapping in that particular context. This process is repeated until all the text in Sindhi Devanagari script is converted into Sindhi Perso-Arabic script. Once done, the system produces the final output text.

## V. EVALUATION

We tested our system for 1500 sentences which had 15497 words and 61993 characters. Among them the rule based system was able to correctly identify 50317 mappings. This gave us an accuracy of 81.17% which was calculated using equation 5. The remaining 11676 characters had ambiguity which were sent to the machine (statistical) learning module for disambiguation. It identified 11463 mappings correctly. This produced an individual ML model's accuracy of 98.18% and the overall accuracy of 99.66%. The system was not able to correctly identify 213 character mapping which was 0.34%. The statistics of this evaluation study is shown in table 3 and table 4.

$$Accuracy = \frac{Correct\ Mappings}{Total\ Characters} \quad (5)$$

TABLE III. SUMMARY OF CORPUS

| Total Sentences | 1500 |
|---|---|
| Total Words | 15497 |
| Total Characters | 61993 |

TABLE IV. SUMMARY OF EVALUATION

| Model | Correct Mappings | Accuracy |
|---|---|---|
| Rule-Based | 50317 | 81.17% |
| Machine Learning | 11463 | 98.18% |
| Overall | 61780 | 99.66% |
| Error | 213 | 0.34% |

## VI. CONCLUSION

In this paper we have shown the working of a transliteration system for Sindhi scripts. For this we have developed a hybrid model which first transcribes Sindhi-Devanagari text into

Sindhi-Perso Arabic text using a rule base. In case if any ambiguity arises where a Devnagari character has multiple mappings in Perso Arabic scripts then an ambiguity resolution module is called. This resolution module is based on probabilistic reasoning. While testing the system it was found that the rule based system was able to correctly transcribe text with an accuracy of 81.17% and the probabilistic model was able to resolve ambiguities with 98.18% accuracy. Overall, the system gave an accuracy of 99.66%.

ACKNOWLEDGMENT

This work is supported by the funding received from SERB, GoI through grant number SPG/2021/003306 for project entitled, "Development of Sindhi-Hindi and Hindi-Sindhi Machine Assisted Translation System".

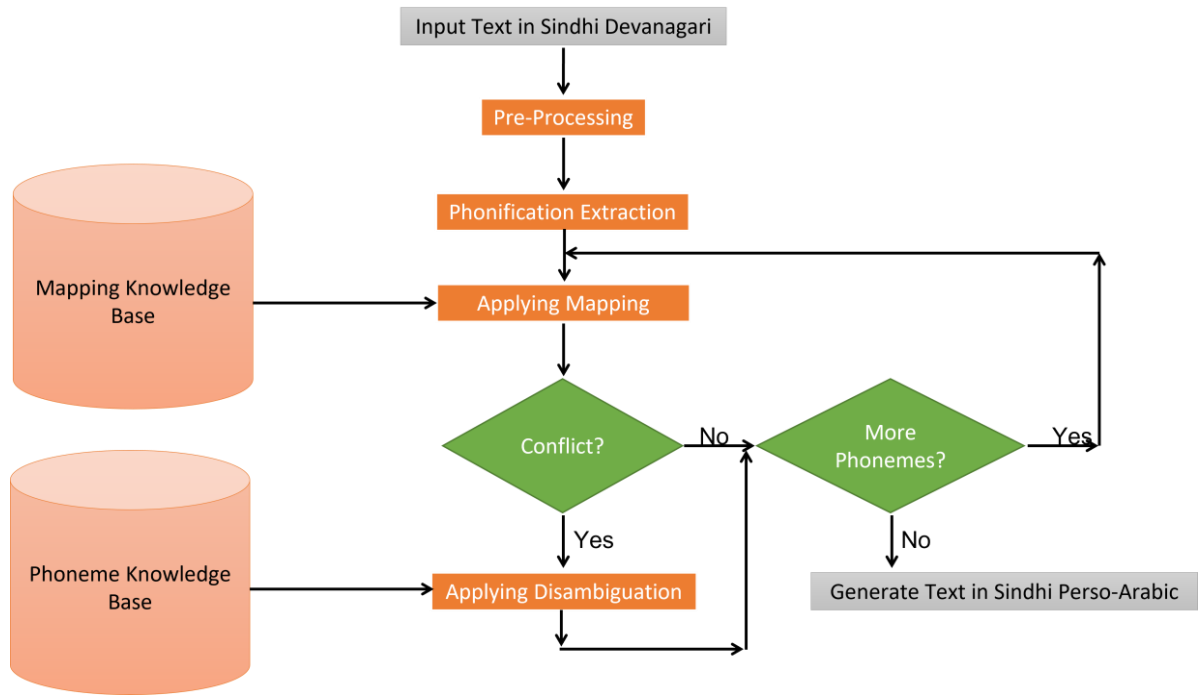

FIGURE 1: WORKING OF SINDHI DEVANAGARI TO SINDHI PERSO-ARABIC MACHINE TRANSLITERATION SYSTEM